\pdfoutput=1

\documentclass[11pt]{article}

\usepackage[]{acl}

\usepackage{times}
\usepackage{latexsym}

\usepackage[T1]{fontenc}

\usepackage[utf8]{inputenc}

\usepackage{microtype}

\usepackage[flushleft]{threeparttable}
\usepackage{pifont}
\usepackage{dsfont}
\usepackage{booktabs,caption}
\usepackage{times}
\usepackage{latexsym}
\usepackage{color}
\usepackage{multirow}
\usepackage{forest}
\usepackage{tikz}
\usepackage{algorithm}
\usepackage{algpseudocode}
\usepackage{amsmath}
\usepackage[encapsulated]{CJK}
\usepackage{graphics}
\usepackage{graphicx}
\usepackage{mwe}
\usepackage{amssymb}
\usepackage{bm}
\usepackage{amsmath}
\usepackage{makecell}
\usepackage{tabularx}
\usepackage{arydshln}
\usepackage[capitalize]{cleveref}
\usepackage{mathrsfs}
\usepackage{arydshln}
\crefformat{section}{\S#2#1#3} 
\usepackage{url}
\usepackage{CJKutf8}
\usepackage{wrapfig}
\usepackage{enumitem}
\usepackage{diagbox}
\usepackage{makecell}
\usepackage{caption}
\usepackage{subcaption}
\usetikzlibrary{calc}
\definecolor{hzw}{RGB}{223, 97, 76}

\definecolor{xing}{RGB}{18, 21, 141}

\title{Bridging the Data Gap between Training and Inference \\ for Unsupervised Neural Machine Translation}

\author{
Zhiwei He\thanks{\ \ Work was done when Zhiwei He was interning at Tencent AI Lab.}\\Shanghai Jiao Tong University\\ \normalsize \sf zwhe.cs@sjtu.edu.cn \And
Xing Wang\\Tencent AI Lab\\ \normalsize \sf brightxwang@tencent.com \AND
Rui Wang\thanks{\ \ Rui Wang is the corresponding author.}\\Shanghai Jiao Tong University\\ \normalsize \sf wangrui12@sjtu.edu.cn \And
Shuming Shi\\Tencent AI Lab\\ \normalsize \sf shumingshi@tencent.com \And
Zhaopeng Tu\\Tencent AI Lab\\ \normalsize \sf zptu@tencent.com
}

\begin{document}
\maketitle
\begin{abstract}
Back-translation is a critical component of Unsupervised Neural Machine Translation (UNMT), which generates pseudo parallel data from target monolingual data. A UNMT model is trained on the pseudo parallel data with {\bf translated source}, and translates {\bf natural source} sentences in inference. The source discrepancy between training and inference hinders the translation performance of UNMT models. By carefully designing experiments, we identify two representative characteristics of the data gap in source: (1) {\em style gap} (i.e., translated vs. natural text style) that leads to poor generalization capability; (2) {\em content gap} that induces the model to produce hallucination content biased towards the target language.
To narrow the data gap, we propose an online self-training approach, which simultaneously uses the pseudo parallel data $\{$natural source, translated target$\}$ to mimic the inference scenario.
Experimental results on several widely-used language pairs show that our approach outperforms two strong baselines (XLM and MASS) by remedying the style and content gaps.~\footnote{\ Code, data, and trained models are available at \url{https://github.com/zwhe99/SelfTraining4UNMT}.}

\end{abstract}

\section{Introduction}
In recent years, there has been a growing interest in unsupervised neural machine translation (UNMT), which requires only monolingual corpora to accomplish the translation task ~\citep{lample2018unsupervised,lample2018phrase,artetxe2018unsupervised,yang2018unsupervised,ren2019unsupervised}.
The key idea of UNMT is to use back-translation (BT) ~\cite{sennrich-etal-2016-improving} to construct the pseudo parallel data for translation modeling.
Typically, UNMT back-translates the natural target sentence into the synthetic source sentence (translated source) to form the training data. A BT loss is calculated on the pseudo parallel data $\{$translated source, natural target$\}$ to update the parameters of UNMT models. 

\begin{table}[t]
    \centering
    \begin{tabular}{c c c}
    \toprule
     & \bf Source & \bf Target\\
    \midrule
    Train &        ~$\mathcal{X^{*}}$ & $\mathcal{Y}$ \\
    Inference &    $\mathcal{X}$       &  ~$\mathcal{Y^{*}}$\\
    \bottomrule
    \end{tabular}
    \caption{ $\left\{\mathcal{X^{*}}, \mathcal{Y}\right\}$ is the translated pseudo parallel data which is used for UNMT training on $X\Rightarrow Y$ translation. The input discrepancy between training and inference: 1) Style gap: $\mathcal{X^*}$  is in translated style, and $\mathcal{X}$ is in the natural style; 2) Content gap: the content of $\mathcal{X^*}$ biases towards target language $Y$ due to the back-translation manipulation, and the content of $\mathcal{X}$ biases towards source language $X$.}
    \label{tab:unmt_data_gap}
\end{table}

In Supervised Neural Machine Translation (SNMT), ~\citet{Edunov:2020:ACL} found that BT suffers from the translationese problem~\cite{Zhang:2019:WMT,Graham:2020:EMNLP} in which BT improves BLEU score on the target-original test set with limited gains on the source-original test set. Unlike authentic parallel data available in the SNMT training data, the UNMT training data entirely comes from pseudo parallel data generated by the back-translation. Therefore in this work, we first revisit the problem in the UNMT setting and start our research from an observation (\cref{sec:revisit-translationese}): with comparable translation performance on the full test set, the BT based UNMT models achieve better translation performance than the SNMT model on the target-original (i.e. translationese) test set, while achieves worse performance on the source-original ones.   

In addition, the pseudo parallel data $\{$translated source, natural target$\}$ generated by BT poses great challenges for UNMT, as shown in Table~\ref{tab:unmt_data_gap}. First, there exists the input discrepancy between the translated source (translated style) in UNMT training data and the natural source (natural style) in inference data. We find that the poor generalization capability caused by the \emph{style gap} (i.e., translated style v.s natural style) limited the UNMT translation performance (\cref{sec:style_gap}). Second, the translated pseudo parallel data suffers from the language coverage bias problem~\cite{wang2021language}, in which the content of UNMT training data biases towards the target language while the content of the inference data biases towards the source language. 
The \emph{content gap} results in hallucinated translations~\cite{lee2018hallucinations,wang2020exposure} biased towards the target language (\cref{sec:content_gap}). 

To alleviate the data gap between the training and inference, we propose an online self-training (ST) approach to improve the UNMT performance.
Specifically, besides the BT loss, the proposed approach also synchronously calculates the ST loss on the pseudo parallel data $\{$natural source, translated target$\}$ generated by self-training to update the parameters of UNMT models.
The pseudo parallel data $\{$natural source, translated target$\}$ is used to mimic the inference scenario with $\{$natural source, translated target$\}$ to bridge the data gap for UNMT.
It is worth noting that the proposed approach does not cost extra computation to generate the pseudo parallel data $\{$natural source, translated target$\}$\footnote{The vanilla UNMT model adopts the dual structure to train both translation directions together, and the pseudo parallel data $\{$natural source,  translated target$\}$ has already been generated and is used to update the parameters of UNMT model in the reverse direction.}, which makes the proposed method efficient and easy to implement. 

We conduct experiments on the XLM~\cite{lample2019cross} and MASS~\cite{song2019mass} UNMT models on multiple language pairs with varying corpus sizes (WMT14 En-Fr / WMT16 En-De / WMT16 En-Ro / WMT20 En-De / WMT21 En-De).
Experimental results show that the proposed approach achieves consistent improvement over the baseline models. Moreover, we conduct extensive analyses to understand the proposed approach better, and the quantitative evidence reveals that the proposed approach narrows the style and content gaps to achieve the improvements. 

In summary, the contributions of this work are detailed as follows:
\begin{itemize}[leftmargin=10pt]
\item Our empirical study demonstrates that the back-translation based UNMT framework suffers from the translationese problem, causing the inaccurate evaluation of UNMT models on standard benchmarks. 
 
\item We empirically analyze the data gap between training and inference for UNMT and identify two critical factors: style gap and content gap.  

\item We propose a simple and effective approach for incorporating the self-training method into the UNMT framework to remedy the data gap between the training and inference. 

\end{itemize}

\section{Translationese Problem in UNMT}
\label{sec:revisit-translationese}
\subsection{Background: UNMT}
\paragraph{Notations.} 
Let $X$ and $Y$ denote the language pair, and let $\mathcal{X}=\{x_{i}\}_{i=1}^M$ and $\mathcal{Y}=\{y_{j}\}_{j=1}^N$ represent the collection of monolingual sentences of the corresponding language, where $M,N$ are the size of the corresponding set.
Generally, UNMT method that based on BT adopts dual structure to train a bidirectional translation model~\cite{artetxe2018unsupervised,artetxe2019effective,lample2018unsupervised,lample2018phrase}.
For the sake of simplicity, we only consider translation direction $X\rightarrow Y$ unless otherwise stated. 

\paragraph{Online BT.} 
Current mainstream of UNMT methods turn the unsupervised task into the synthetic supervised task through BT, which is the most critical component in UNMT training. Given the translation task $X \rightarrow Y$ where target corpus $\mathcal{Y}$ is available, 
for each batch, the target sentence $y \in \mathcal{Y}$ is used to generate its synthetic source sentence by the backward model MT$_{Y \rightarrow X}$:

\begin{equation}
    x^{*} = \mathop{\arg\max}_{x} P_{Y\rightarrow X}(x\mid y;\tilde{\theta}),
\label{eq:bt_generation}
\end{equation}
where $\tilde{\theta}$ is a fixed copy of the current parameters $\theta$ indicating that the gradient is not propagated through $\tilde{\theta}$. In this way, the synthetic parallel sentence pair $\{x^{*},y\}$ is obtained and used to train the forward model MT$_{X \rightarrow Y}$ in a supervised manner by minimizing:
\begin{equation}
    \mathcal{L}_{B} = \mathbb{E}_{y\sim\mathcal{Y}} [-\log{P_{X \rightarrow Y}(y\mid x^{*};\theta)}]. 
\label{eq:back_tralsation_loss}
\end{equation}

It is worth noting that the synthetic sentence pair generated by the BT is the only supervision signal of UNMT training.

\paragraph{Objective function.} In addition to BT, denoising auto-encoding (DAE) is an additional loss term of UNMT training, which is denoted by $\mathcal{L}_{D}$ and is not the main topic discussed in this work.

In all, the final objective function of UNMT is:
\begin{equation}
   \mathcal{L} = \mathcal{L}_{B} + \lambda_{D}\mathcal{L}_{D},
\label{unmt-final-obj}
\end{equation}
where $\lambda_{D}$ is the hyper-parameter weighting DAE loss term. 
Generally, $\lambda_{D}$ starts from one and decreases as the training procedure continues\footnote{Verified from open-source XLM Github implementation.}.

\subsection{Translationese Problem}
\label{sec:translationese-problem}
To verify whether the UNMT model suffers from the input gap between training and inference and thus is biased towards translated input while against natural input, we conduct comparative experiments between SNMT and UNMT models.

\paragraph{Setup}
We evaluate the UNMT and SNMT models on WMT14 En-Fr, WMT16 En-De and WMT16 En-Ro test sets, following~\citet{lample2019cross} and~\citet{song2019mass}. 
We first train the UNMT models on the above language pairs with model parameters initialized by XLM and MASS models.
Then, we train the corresponding SNMT models whose performance on the full test sets is controlled to be approximated to UNMT by undersampling training data.
Finally, we evaluate the UNMT and SNMT models on the target-original and source-original test sets, whose inputs are translated and natural respectively.
Unless otherwise stated, we follow previous work~\cite{lample2019cross,song2019mass} to use case-sensitive BLEU score \citep{papineni2002bleu} with the \texttt{multi-bleu.perl}\footnote{\url{https://github.com/moses-smt/mosesdecoder/blob/master/scripts/generic/multi-bleu.perl}} script as the evaluation metric.
Please refer to \cref{sec:sacrebleu-results} for the results of SacreBLEU, and refer to \cref{sec:training-detail} for the training details of SNMT and UNMT models.

\begin{table}[t]  
    \centering
    \setlength\tabcolsep{3.3pt}
    \begin{tabular}{c rr rr rr c}
        \toprule
        \multirow{2}{*}{\bf Model}& 
        \multicolumn{2}{c}{\bf En-Fr} & 
        \multicolumn{2}{c}{\bf En-De} &  
        \multicolumn{2}{c}{\bf En-Ro} &    
        \multirow{2}{*}{\bf Avg.}\\ 
        \cmidrule(lr){2-3} \cmidrule(lr){4-5} \cmidrule(lr){6-7} 
        & $\Rightarrow$~ & 
          $\Leftarrow$~  &  
          $\Rightarrow$~ & 
          $\Leftarrow$~  & 
          $\Rightarrow$~ & 
          $\Leftarrow$~ \\ 
          
    \midrule
    \multicolumn{8}{l}{\bf Full Test Set}\\
    SNMT & 38.4 & 33.6 & 29.5 & 33.9 & 33.7 & 32.5 & 33.6\\
    \hdashline
    XLM   & 37.4 & 34.5 & 27.2 & 34.3 & 34.6 & 32.7 & 33.5\\
    MASS  & 37.8 & 34.9 & 27.1 & 35.2 & 35.1 & 33.4 & 33.9\\
    \midrule
    \multicolumn{8}{l}{\bf Target-Original Test Set / Translated Input}\\
    SNMT & 37.4 & 32.4 & 25.6 & 37.1 & 38.2 & 28.2 & 33.2\\
    \hdashline
    XLM  & \bf39.1 & \bf36.5 & \bf26.6 & \bf42.2 & \bf42.1 & \bf34.4 & \bf36.8\\
    MASS & \bf39.2 & \bf37.6 & \bf27.0 & \bf42.9 & \bf43.1 & \bf35.6 & \bf37.6\\
    \midrule
    \multicolumn{8}{l}{\bf Source-Original Test Set / Natural Input}\\
    SNMT & \bf  38.2 & \bf34.1 & \bf32.3 & \bf28.8 & \bf29.4 & \bf35.9 & \bf33.1\\
    \hdashline
    XLM  &    34.7 &    30.4 &    26.6 &    22.5 &    27.4 &    30.6 &    28.7\\
    MASS &    35.2 &    30.2 &    26.1 &    23.6 &    27.4 &    30.8 &   28.9\\
    \bottomrule
    \end{tabular}
\caption{Translation performance of SNMT and UNMT models on full / target-original / source-original test sets. SNMT denotes the supervised translation models trained on undersampled parallel data and their performance on full test data are controlled to be approximate to the UNMT counterparts.}
\label{tab:snmt-unmt-translationese}
\end{table}

\paragraph{Results}
We present the translation performance in terms of the BLEU score in Table~\ref{tab:snmt-unmt-translationese} and our observations are:

\begin{itemize}[leftmargin=10pt]
\item UNMT models perform close to the SNMT models on the full test sets with 0.3 BLEU difference at most on average (33.5/33.9 vs. 33.6).

\item UNMT models outperform SNMT models on target-original test sets (translated input) with average BLEU score improvements of 3.6 and 4.4 BLEU points (36.8/37.6 vs. 33.2).

\item  UNMT models underperform the SNMT models on source-original test sets (natural input) with an average performance degradation of 4.4 and 4.2 BLUE points (28.7/28.9 vs. 33.1).
\end{itemize}

The above observations are invariant concerning the pre-trained model and translation direction. 
In particular, the unsatisfactory performance of UNMT under natural input indicates that UNMT is overestimated on the previous benchmark.
We attribute the phenomenon to the data gap between training and inference for UNMT: there is a mismatch between natural inputs of source-original test data and the back-translated inputs that UNMT employed for training.
This work focuses on the experiments on the source-original test sets (i.e., the input of an NMT translation system is generally natural), which is closer to the practical scenario.\footnote{From WMT19, the WMT community proposes to use the source-original test with natural input sets to evaluate the translation performance.}  

\section{Data Gap between Training and Inference}
In this section, we identity two representative data gaps between training and inference data for UNMT: style gap and content data. 
We divide the test sets into two portions: the natural input portion with source sentences originally written in the source language and the translated input portion with source sentences translated from the target language. 
Due to the limited space, we conduct the experiments with pre-trained XLM initialization and perform analysis with different kinds of inputs (i.e., natural and translated inputs) on De$\Rightarrow$En newstest2013-2018 unless otherwise stated.
\label{sec:data-gap-between-training-and-inference-of-unmt}

\subsection{Style Gap}
\label{sec:style_gap} 

\begin{table}[t]
    \centering
    \begin{tabular}{c c c}
    \toprule
     \bf Inference Input  & \bf PPL  \\
    \midrule
           Natural       & 242 \\
           Translated    & 219 \\
    \bottomrule
    \end{tabular}
    \caption{Perplexity on the natural input sentences and translated input sentences of newstest2013-2018. The language model is trained on the UNMT translated source sentences.}
    \label{tab:style_gap}
\end{table}

To perform the quantitative analysis of the style gap, we adopt KenLM\footnote{https://github.com/kpu/kenlm} to train a 4-gram language model on the UNMT translated source sentences\footnote{To alleviate the content bias problem, we generate the training data 50\% from En$\Rightarrow$De translation and 50\% from round trip translation De$\Rightarrow$En$\Rightarrow$De.} and use the language model to calculate the perplexity (PPL) of natural and translated input sentences in the test sets. 
The experimental results are shown in Table~\ref{tab:style_gap}. 
The lower perplexity value (219 < 242) indicates that \textbf{compared with the natural inputs, the UNMT translated training inputs have a more similar style with translated inputs in the test sets}.
\begin{table}[t]  
    \centering
    \setlength\tabcolsep{3.0pt}
    \begin{tabular}{c c c c c}
        \toprule
    \multirow{2}{*}{\bf Model} & \multicolumn{2}{c}{\bf Natural De} & \multicolumn{2}{c}{\bf Translated De$^{*}$} \\
    \cmidrule(lr){2-3} \cmidrule(lr){4-5}  
     & BLEU & $\Delta$ & BLEU & $\Delta$\\ 
    \midrule
    SNMT  & 28.8  & -- & 44.9 & -- \\
    \hdashline
    UNMT   & 22.5 & -6.3 & 42.1 & -2.8\\
    \bottomrule
    \end{tabular}
\caption{Translation performance on natural input portion of WMT16 De$\Rightarrow$En. We also use Google Translator to generate the translated version by translating the corresponding target sentences. }
\label{tab:natural-src-vs-translated-src}
\end{table}

In order to further reveal the influence of the style gap on UNMT, we manually eliminated it and re-evaluated the models on the natural input portion of WMT16 De$\Rightarrow$En. 
Concretely, We first take the third-party Google Translator to translate the target English sentences of the test sets into the source German language to eliminate the style gap. 
And then we conduct translation experiments on the natural input portion and its Google translated portion to evaluate the impact of the style gap on the translation performance. 
We list the experimental results in Table~\ref{tab:natural-src-vs-translated-src}. 
We can find that by converting from the natural inputs (natural De) to the translated inputs (translated De$^{*}$), the UNMT model achieves more improvement than the SNMT model (-2.8 > -6.3), demonstrating that the style gap inhibits the UNMT translation output quality.

\subsection{Content Gap}
\label{sec:content_gap}
In this section, we show the existence of the content gap by (1) showing the most high-frequency name entities,  (2) calculating content similarity using term frequency-inverse document frequency (TF-IDF) for the training and inference data.

We use spaCy\footnote{https://github.com/explosion/spaCy} to recognize German named entities for the UNMT translated source sentences, natural inputs and translated inputs in test sets, and show the ten most frequent name entities in Table~\ref{tab:top_words}. From the table, we can observe that the UNMT translated source sentences have few named entities biased towards source language German (words in red color), while having more named entities biased towards target language English, e.g., USA, Obama. It indicates that the content of the UNMT translated source sentences is biased towards the target language English. 

Meanwhile, the natural input portion of the inference data has more named entities biased towards source language German (words in red color), demonstrating that the content gap exists between the natural input portion of the inference data and the UNMT translated training data.

\begin{table}[t]
    \centering
    \setlength{\tabcolsep}{3.5pt}
\small{
    \begin{tabular}{c l}
    \toprule
    \bf Data & \bf Most Frequent Name Entities  \\
    \midrule
    \multirow{2}{*}{\shortstack{Natural\\Infer. Input}}  &  \small{\textcolor{red}{Deutschland}, Stadt, \textcolor{red}{CDU, deutschen}, Zeit} \\
     & \small{\textcolor{red}{SPD}, USA, \textcolor{red}{deutsche}, China, Mittwoch}\\
      \hdashline
    \multirow{3}{*}{\shortstack{Translated\\Infer. Input}} & \small{\textcolor{blue}{Großbritannien, London, Trump, USA}, }\\
        & \small{Russland, \textcolor{blue}{Vereinigten Staaten, Europa}}\\
        & \small{Mexiko, \textcolor{blue}{Amerikaner, Obama}}\\
    \midrule
    \multirow{2}{*}{\shortstack{BT\\Train Data}}   &  \small{\textcolor{red}{Deutschland}, \textcolor{red}{dpa}, \textcolor{blue}{USA}, China, \textcolor{blue}{Obama}, Stadt} \\
       & \small{Hause, \textcolor{blue}{Europa}, \textcolor{blue}{Großbritannien}, Russland} \\
    \bottomrule
    \end{tabular}
}
    \caption{Ten most frequent entities in the source sentences (i.e., German) of back-translated training data (``BT Train Data"). For reference, we also list the most frequent entities in the natural and translated inference inputs.
    The BT training data has more entities biased towards the target language English (\textcolor{blue}{blue words}) rather than the expected source language German (\textcolor{red}{red words}).}
    \label{tab:top_words}
\end{table}

\begin{table}[t]
    \centering
    \begin{tabular}{c c c}
    \toprule           
    \multirow{2}{*}{\bf Inference Input} &                    \multicolumn{2}{c}{\bf Train}  \\ 
     \cmidrule(lr){2-3}
      & Natural &  Translated    \\
    \midrule
        Natural      &   0.95  & 0.85  \\
        Translated  &   0.84  & 0.93  \\
       
    \bottomrule
    \end{tabular}
    \caption{Content similarity between different kinds of training and inference data.}

    \label{tab:content-gap-similarity}
\end{table}

Next, we remove the stop words and use the term frequency-inverse document frequency (TF-IDF) approach to calculate the content similarity between the training and inference data.
Similarity scores are presented in Table~\ref{tab:content-gap-similarity}.
We can observe that the UNMT translated source data has a more significant similarity score with translated inputs which are generated from the target English sentences.
This result indicates that \textbf{the content of UNMT translated source data is more biased towards the target language}, which is consistent with the findings in Table~\ref{tab:top_words}.

\begin{table}[t]
    \centering
    \begin{tabular}{c l}
    \toprule
    \multirow{2}{*}{ Input} &  \small{Die \textcolor{red}{deutschen} Kohlekraftwerke ... der in }  \\
                               &  \small{ \textcolor{red}{Deutschland} emittierten Gesamtmenge .}  \\
    \midrule
   \multirow{2}{*}{Ref} &  \small{\textcolor{red}{German} coal plants , ..., two thirds of} \\
                              & \small{ the total amount emitted in \textcolor{red}{Germany} .} \\
    \midrule
    \multirow{2}{*}{SNMT}     &  \small{..., \textcolor{red}{German} coal-fired power stations ...} \\
                              & \small{of the total emissions in \textcolor{red}{Germany} .} \\
   \midrule
   \multirow{2}{*}{UNMT}     &   \small{\textcolor{blue}{U.S.} coal-fired power plants ... two thirds of}  \\
                             &   \small{the total amount emitted in the \textcolor{blue}{U.S.} ... .}\\
    \bottomrule
    \end{tabular}
    \caption{Example translation that the UNMT model outputs the hallucinated translation ``U.S.'', which is biased towards target language English. }
    \label{tab:content_gap}
\end{table}

As it is difficult to measure the name entities translation accuracy in terms of BLEU evaluation metric, we provide a translation example in Table~\ref{tab:content_gap} to show the effect of the content gap in the UNMT translations (more examples in \cref{sec:tranlstion_examples}). We observe that the UNMT model outputs the hallucinated translation ``U.S.'', which is biased towards the target language English. We present a quantitative analysis to show the impact of the content gap on UNMT translation performance in Section~\ref{sec:anaysis_data_gap}.

\section{Online Self-training for UMMT}
To bridge the data gap between training and inference of UNMT, we propose a simple and effective method through self-training. 
For the translation task $X\rightarrow Y$, we generate the source-original training samples from the source corpus $\mathcal{X}$ to improve the model's translation performance on natural inputs.
For each batch, we apply the forward model MT$_{X \rightarrow Y}$ on the natural source sentence $x$ to generate its translation:
\begin{equation}
    y^{*} = \mathop{\arg\max}_{y} P_{X\rightarrow Y}(y\mid x;\tilde{\theta}).
\label{eq:st_generation}
\end{equation}
In this way, we build a sample $\{x, y^*\}$ with natural input, on which the model can be trained by minimizing:
\begin{equation}
    \mathcal{L}_{S} = \mathbb{E}_{x\sim\mathcal{X}} [-\log{P_{X \rightarrow Y}(y^{*}\mid x;\theta)}]. 
\label{eq:self_training_loss}
\end{equation}
Under the framework of UNMT training, the final objective function can be formulated as:
\begin{equation}
   \mathcal{L} = \mathcal{L}_{B} + \lambda_{D}\mathcal{L}_{D}+ \lambda_{S}\mathcal{L}_{S},
\label{unmt-st-final-obj}
\end{equation}
where $\lambda_S$ is the hyper-parameter weighting the self-training loss term.
It is worth noting that the generation step of Eq.\eqref{eq:st_generation} has been done by the BT step of $Y\rightarrow X$ training. 
Thus, the proposed method will not increase the training cost significantly but make the most of the data generated by BT (\tablename~\ref{tab:wmt19_20}).

\section{Experiments}
\label{sec:experiemnts}

\subsection{Setup}
\label{sec:experiemnts-setup}
\paragraph{Data}
We follow the common practices to conduct experiments on several UNMT benchmarks: WMT14 En-Fr, WMT16 En-De, WMT16 En-Ro.
The details of monolingual training data are delineated in~\cref{para:training-data-for-unmt}.
We adopt En-Fr newsdev2014, En-De newsdev2016, En-Ro newsdev2016 as the validation (development) sets, and En-Fr newstest2014, En-De newstest2016, En-Ro newstest2016 as the test sets.
In addition to the full test set, we split the test set into two parts: target-original and source-original, and evaluate the model's performance on the three kinds of test sets.
We use the released XLM BPE codes and vocabulary for all language pairs.

\begin{table*}[htpb]  
    \centering
    \begin{tabular}{l l l cc cc cc cc}
        \toprule
        \multirow{2}{*}{\bf Testset} &  \multirow{2}{*}{\bf Model} &    \multirow{2}{*}{\bf Approach}  &  
        \multicolumn{2}{c}{\bf En-Fr} & 
        \multicolumn{2}{c}{\bf En-De} &  
        \multicolumn{2}{c}{\bf En-Ro} &    
        \multirow{2}{*}{\bf Avg.}  &  \multirow{2}{*}{\bf $\Delta$}\\ 
        
        & & & $\Rightarrow$~ & 
          $\Leftarrow$~  &  
          $\Rightarrow$~ & 
          $\Leftarrow$~  & 
          $\Rightarrow$~ & 
          $\Leftarrow$~ \\ 

    \specialrule{.05em}{.1ex}{.1ex}
    \multicolumn{11}{c}{\textit{Existing Works (Full set)}}   \\
    \specialrule{.05em}{.1ex}{.1ex}
    \multicolumn{3}{l}{XLM~\cite{lample2019cross}}         & 33.4 & 33.3 & 26.4 & 34.3 & 33.3 & 31.8  & 32.1 & --\\
    \multicolumn{3}{l}{MASS~\cite{song2019mass}}           & 37.5 & 34.9 & 28.3 & 35.2 & 35.2 & 33.1  & 34.0 & --\\
    \multicolumn{3}{l}{CBD ~\cite{nguyen2021cbd}}          & 38.2 & 35.5 & 30.1 & 36.3 & 36.3 & 33.8  & 35.0 & --\\

    \specialrule{.05em}{.1ex}{.1ex}
    \multicolumn{11}{c}{\textit{Our Implementation}}   \\
    \specialrule{.05em}{.1ex}{.1ex}
    \multirow{4}{*}{\bf Full set} & \multirow{2}{*}{XLM}    & UNMT    &    37.4 &    34.5 &    27.2 &    34.3 &    34.6 &    32.7  &    33.5 & --\\
    & & ~~+Self-training                                              & \bf37.8 & \bf35.1 & \bf28.1 & \bf34.8 & \bf36.2 & \bf33.9  & \bf34.3 & +0.8\\
     \cline{2-11}
    &  \multirow{2}{*}{MASS}    & UNMT                                &    37.8 &    34.9 &    27.1 &    35.2 &    35.1 &    33.4  &    33.9 & --\\
     & & ~~+Self-training                                             & \bf38.0 & \bf35.2 & \bf28.9 & \bf35.6 & \bf36.5 & \bf34.0  & \bf34.7 & +0.8\\
    \midrule
    \multirow{4}{*}{\bf Trg-Ori}  &  \multirow{2}{*}{XLM}  & UNMT     &    39.1 &    36.5 & \bf26.6 &    42.2 &    42.1 & \bf34.4  &    36.8 & --\\
    & & ~~+Self-training                                              & \bf39.3 & \bf37.8 &    26.5 & \bf42.4 & \bf42.9 &    34.1  & \bf37.2 & +0.4\\
     \cline{2-11}
     &  \multirow{2}{*}{MASS}                             & UNMT      & \bf39.2 & \bf37.6 &    27.0 & \bf42.9 & \bf43.1 & \bf35.6  & \bf37.6 & --\\
     & & ~~+Self-training                                             &    39.0 &    37.3 & \bf27.7 &    42.7 &    42.9 &    35.3  &    37.5 & -0.1\\
    \midrule
    \multirow{4}{*}{\bf Src-Ori}  &  \multirow{2}{*}{XLM}  & UNMT     &    34.7 & \bf30.4 &    26.6 &    22.5 &    27.4 &    30.6  &    28.7 & --\\
    & & ~~+Self-training                                              & \bf35.4$^\Uparrow$ &    30.2 & \bf28.0$^\Uparrow$ & \bf23.1$^\uparrow$ & \bf29.6$^\Uparrow$ & \bf32.7$^\Uparrow$  & \bf29.8 & +1.1\\
     \cline{2-11}
    &  \multirow{2}{*}{MASS}    & UNMT                                &    35.2 &    30.2 &    26.1 &    23.6 &    27.4 &    30.8  &    28.9 & --\\
    & & ~~+Self-training                                              & \bf35.9$^\Uparrow$ & \bf30.9$^\uparrow$ & \bf28.7$^\Uparrow$ & \bf24.9$^\Uparrow$ & \bf30.1$^\Uparrow$ & \bf31.9$^\Uparrow$  & \bf30.4 & +1.5\\
    \bottomrule
    \end{tabular}
    \caption{Translation performance on WMT14 En-Fr, WMT16 En-De, WMT16 En-Ro and their corresponding source-original (natural input) and target-original (translated input) subset. ``$\uparrow/\Uparrow$'': significant over the corresponding baseline model ($p < 0.05/0.01$), tested by bootstrap resampling~\cite{Koehn2004:emnlp}.}
    \label{tab:main-result}
\end{table*}

\paragraph{Model}
We evaluate the UNMT model fine-tuned on XLM\footnote{https://github.com/facebookresearch/XLM\label{fn:xlm-code}} and MASS\footnote{https://github.com/microsoft/MASS\label{fn:mass-code}} pre-trained model~\citep{lample2019cross,song2019mass}.
For XLM models, we adopt the pre-trained models released by \citet{lample2019cross} for all language pairs.
For MASS models, we adopt the pre-trained models released by \citet{song2019mass} for En-Fr and En-Ro and continue pre-training the MASS model of En-De for better reproducing the results. 
More details are delineated in~\cref{sec:training-detail-unmt}.

\subsection{Main Result}
\label{sec:main-result}
\tablename~\ref{tab:main-result} shows the translation performance of XLM and MASS baselines and our proposed models. We have the following observations:
\begin{itemize}[leftmargin=10pt]
\item  Our re-implemented baseline models achieve comparable or even better performance as reported in previous works.
The reproduced XLM+UNMT model has an average improvement of 1.4 BLEU points compared to the original report in~\citet{lample2019cross} and MASS+UNMT model is only 0.1 BLEU lower on average than~\citet{song2019mass}.

\item  Our approach with online self-training significantly improves overall translation performance (+0.8 BLEU on average).
This demonstrates the universality of the proposed approach on both large-scale (En-Fr, En-De) and data imbalanced corpus (En-Ro).

\item  In the translated input scenario, our approach achieves comparable performance to baselines. It demonstrates that although the sample of self-training is source-original style, our approach does not sacrifice the performance on the target-original side.

\item  In the natural input scenario, we find that our proposed approach achieves more significant improvements, with +1.1 and +1.3 average BLEU on both baselines. The reason is that the source-original style sample introduced by self-training alleviates model bias between natural and translated input.
\end{itemize}

\begin{table*}[htpb]   
\centering
    \begin{tabular}{l l ll  ll cc c}
        \toprule
        \multirow{2}{*}{\bf Model}     &    \multirow{2}{*}{\bf Approach} & 
        \multicolumn{2}{c}{\bf WMT19} & 
        \multicolumn{2}{c}{\bf WMT20} & 
        \multirow{2}{*}{\bf Avg.}    &  \multirow{2}{*}{\bf $\Delta$} &  \multirow{2}{*}{\bf Training Cost}\\
    
        \cmidrule(lr){3-4} \cmidrule(lr){5-6}
    
        & & \bf ~~$\Rightarrow$ & 
          \bf ~~$\Leftarrow$  &  
          \bf ~~$\Rightarrow$ & 
          \bf ~~$\Leftarrow$  \\ 
        \midrule

   \multirow{4}{*}{XLM}    & UNMT &    26.6 &    24.4 &    22.9 &    26.6 &    25.1 & -- & 1.0 \\
    & ~~+Offline ST             & 26.9 & 24.2 & 23.2 & 25.9 & 25.1 & +0.0 & $\times$1.8 \\
   & ~~+CBD &  28.3$^\Uparrow$ & 25.6$^\Uparrow$ & 24.2$^\Uparrow$ & 26.9 & 26.3 & +1.2 & $\times$7.3\\
  
   & ~~+Online ST             & \bf28.3$^\Uparrow$ & \bf26.0$^\Uparrow$ & \bf24.3$^\Uparrow$ & \bf27.6$^\Uparrow$ & \bf26.6 & +1.5 & $\times$1.2\\
    
    \midrule
   \multirow{4}{*}{MASS}  & UNMT &    26.7 &    24.6 &    23.1 &    27.0 &    25.3 & -- & 1.0 \\
   & ~~+Offline ST             & 27.2 & 24.6 & 23.1 & 26.9 & 25.4  & +0.1 & $\times$1.8 \\ 
    & ~~+CBD  & 28.3$^\Uparrow$ & 25.6$^\Uparrow$ & \bf24.0$^\Uparrow$ & 27.0 & 26.2 & +0.9 & $\times$7.3\\
    & ~~+Online ST           & \bf28.5$^\Uparrow$ & \bf26.1$^\Uparrow$ & 23.8$^\Uparrow$ & \bf27.8$^\Uparrow$ & \bf26.6 & +1.3 & $\times$1.1 \\
    \bottomrule
    \end{tabular}
    \caption[Caption for LOF]{Comparison with offline self-training and CBD\protect\footnotemark. ``$\uparrow/\Uparrow$'': significant over the corresponding baseline model ($p < 0.05/0.01$), tested by bootstrap resampling~\cite{Koehn2004:emnlp}. The training cost is estimated by the time required for training one epoch where the cost of data generation is also considered.}
    \label{tab:wmt19_20}
\end{table*}

\subsection{Comparison with Offline Self-training and CBD}
We compare online self-training with the following two related methods, which also incorporate natural inputs in training:
\begin{itemize}
    \item \textbf{Offline Self-training} model distilled from the forward and backward translated data generated by the trained UNMT model.
    \item \textbf{CBD}~\cite{nguyen2021cbd} model distilled from the data generated by two trained UNMT models through cross-translation, which embraces data diversity.
\end{itemize}
\paragraph{Dataset}
Previous studies have recommended restricting test sets to natural input sentences, a methodology adopted by the 2019-2020 edition of the WMT news translation shared task~\cite{Edunov:2020:ACL}.
In order to further verify the effectiveness of the proposed approach, we also conduct the evaluation on WMT19 and WMT20 En-De test sets.
Both test sets contain only natural input samples.

\paragraph{Results}
\footnotetext{Our re-implemented CBD model can not achieve comparable performance with~\citet{nguyen2021cbd}, with 28.4 and 35.2 BLEU scores on WMT16 En-De and De-En test sets.}
Experimental results are presented in Table~\ref{tab:wmt19_20}. We also show the training costs of these methods. We find that
\begin{itemize}[leftmargin=10pt]
\item Unexpectedly, the offline self-training has no significant improvement over baseline UNMT. 
\citet{sun-etal-2021-self} have demonstrated the effectiveness of offline self-training in UNMT under low-resource and data imbalanced scenarios. However, in our data-sufficient scenarios, offline self-training may suffer from the data diversity problem while online self-training can alleviate the problem through the dynamic model parameters during the training process. We leave the complete analysis to future work.

\item CBD achieves a significant improvement compared to baseline UNMT, but the training cost is about six times that of online self-training.

\item The proposed online self-training achieves the best translation performance in terms of BLEU score, which further demonstrates the superiority of the proposed method under natural input.
\end{itemize}

\section{Analysis}

\subsection{Translationese Output}
Since the self-training samples are translated sentences on the target side, there is concern that the improvement achieved by self-training only comes from making the model outputs better match the translated references, rather than enhancing the model's ability on natural inputs.
To dispel the concern, we conducted the following experiments: (1) evaluate the fluency of model outputs in terms of language model PPL and (2) evaluate the translation performance on Google Paraphrased WMT19 En$\Rightarrow$De test sets~\cite{freitag-bleu-paraphrase-references-2020}.
\paragraph{Output fluency}
We exploit the monolingual corpora of target languages to train the 4-gram language models. 
\tablename~\ref{tab:ppl} shows the language models' PPL on model outputs of test sets mentioned in \cref{sec:main-result}.
We find that online self-training has only a slight impact on the fluency of model outputs, with the average PPL of XLM and MASS models only increasing by +3 and +6, respectively. We ascribe this phenomenon to the translated target of self-training samples, which is model generated and thus less fluent then natural sentences. However, since the target of BT data is natural and the BT loss term is the primary training objective, the output fluency does not decrease significantly.

\begin{table}[htpb]   
\centering
\setlength\tabcolsep{3.3pt}
    \begin{tabular}{l cc  cc cc c}
        \toprule
        \multirow{2}{*}{\bf Approach} & 
        \multicolumn{2}{c}{\bf En-Fr} & 
        \multicolumn{2}{c}{\bf En-De} &  
        \multicolumn{2}{c}{\bf En-Ro} & 
        \multirow{2}{*}{\bf Avg.}  \\
        \cmidrule(lr){2-3} \cmidrule(lr){4-5} \cmidrule(lr){6-7}
        & \bf $\Rightarrow$ & 
          \bf $\Leftarrow$  &  
          \bf $\Rightarrow$ & 
          \bf $\Leftarrow$ &  
          \bf $\Rightarrow$ & 
          \bf $\Leftarrow$   \\ 
        \midrule

   \multicolumn{8}{c}{\bf XLM} \\
   UNMT &  101  & 147 & 250 & 145 & 152 & 126 & 154\\
   ~~+ST             &  101  & 144 & 253 & 147 & 156 & 138 & 157 \\
    
    \midrule
    \multicolumn{8}{c}{\bf MASS} \\
    UNMT &  100 & 145 & 256 & 144 & 143 & 119 & 151\\
    ~~+ST             &  103 & 146 & 263 & 142 & 156 & 133 & 157\\
    \bottomrule
    \end{tabular}
    \caption{Automatic fluency analysis  in terms of perplexity (PPL). Language models are trained on the natural monolingual data in the respective target language.}
    \label{tab:ppl}

\end{table}

\paragraph{Translation performance on paraphrased references} 
\citet{freitag-bleu-paraphrase-references-2020} collected additional human translations for newstest2019 with the ultimate aim of generating a natural-to-natural test set. We adopt the HQ(R) and HQ(all 4), which have higher human adequacy rating scores, to re-evaluate our proposed models.

We present the experimental results in \tablename~\ref{tab:google-paraphrase}.
Our proposed method outperforms baselines on both kinds of test sets. Therefore, we demonstrate that our proposed method improves the UNMT model performance on natural input with limited translationese outputs.

\begin{table}[htpb]   
    \setlength\tabcolsep{3.3pt}
    \centering
    \begin{tabular}{l  cc   }
        \toprule
        \multirow{1}{*}{\bf Model}  &  \bf HQ(R) & \bf HQ(all 4) \\
        \midrule
        
    \bf Supervised Model &   \multirow{2}{*}{ 35.0} &  \multirow{2}{*}{ 27.2} \\
    \citep{freitag-bleu-paraphrase-references-2020}   \\
    \midrule
    ~~~XLM+UNMT                      &    24.5 & 19.6\\
    ~~~~~~~+Self-training            & \textbf{25.9} & \textbf{20.7}\\
    \hdashline
    ~~~MASS+UNMT                    &    24.3  & 19.6\\
    ~~~~~~~+Self-training           & \textbf{26.0}  & \textbf{20.8} \\
    \bottomrule
    \end{tabular}
    \caption{Translation performance on WMT19 En$\Rightarrow$De test sets with additional human translation references provided by \citet{freitag-bleu-paraphrase-references-2020}. We report sacreBLEU for comparison with supervised model.}
    \label{tab:google-paraphrase}
\end{table}

\begin{table}[htpb]   
\centering
    \begin{tabular}{l l c}
        \toprule
        \bf Model     &    \bf Approach & 
        \bf NER Acc. \\

\midrule    
   \multirow{2}{*}{XLM} & UNMT  &  0.46 \\
                        & ~~+Self-training  & \bf0.53 \\
\hdashline
   \multirow{2}{*}{MASS} & UNMT  & 0.44 \\
                        & ~~+Self-training      & \bf0.52        \\
    \bottomrule
    \end{tabular}
    \caption{Accuracy of NER translation on natural input portion of test sets.}
    \label{tab:ner-acc}
\end{table}

\subsection{Data Gap}
\label{sec:anaysis_data_gap}
\paragraph{Style Gap} From Table~\ref{tab:main-result}, our proposed approach achieves significant improvements on the natural input portion while not gaining on the translated input portion over the baselines. It indicates our approach has better generalization capability on the natural input portion of test sets than the baselines. 

\paragraph{Content Gap} To verify that our proposed approach bridges the content gap between training and inference, we calculate the accuracy of NER translation by different models. Specifically, we adopt spaCy to recognize the name entities in reference and translation outputs and treat the name entities in reference as the ground truth to calculate the accuracy of NER translation.   
We show the results in \tablename~\ref{tab:ner-acc}.
Our proposed method achieves a significant improvement in the translation accuracy of NER compared to the baseline. The result demonstrates that online self-training can help the model pay more attention to the input content rather than being affected by the content of the target language training corpus.

\subsection{Target Quality}
Next, we investigate the impact of target quality on ST. 
We use the SNMT model from \cref{sec:translationese-problem} to generate ST data rather than the current model itself and keep the process of BT unchanged. 
As shown in \tablename~\ref{tab:snmt-unmt-translationese}, the SNMT models perform well on source-original test set and thus yield higher quality target in ST data. 
We denote this variant as ``knowledge distillation (KD)'' and report the performance on WMT19/20 E$\Leftrightarrow$De in \tablename~\ref{tab:target-quality}. 
When target quality gets better, model performance improves significantly, as expected. Therefore, reducing the noise on the target side of the ST data may further improve the performance.
Implementing in an unsupervised manner is left to future work.

\begin{table}[htpb]
    \centering
    \begin{tabular}{c cc cc}
    \toprule
    \multirow{2}{*}{\bf Approach} & \multicolumn{2}{c}{\bf WMT19} & \multicolumn{2}{c}{\bf WMT20}  \\
    \cmidrule(lr){2-3} \cmidrule(lr){4-5}

    & \bf ~$\Rightarrow$ & 
      \bf ~$\Leftarrow$  &  
      \bf ~$\Rightarrow$ & 
      \bf ~$\Leftarrow$  \\ 
      
     \midrule
        \multicolumn{5}{c}{\bf XLM} \\
    UNMT   & 26.6 & 24.4 & 22.9 & 26.6\\
    ~+ST   & 28.3 & 26.0 & 24.3 & 27.6\\
    ~~+KD  & \bf33.8 & \bf31.0 & \bf29.5 & \bf30.6\\
    \hdashline
    \multicolumn{5}{c}{\bf MASS} \\
    UNMT  & 26.7 & 24.6 & 23.1 & 27.0\\
    ~+ST    & 28.5 & 26.1 & 23.8 & 27.8\\
    ~~+KD    & \bf32.9 & \bf31.0 & \bf28.1 & \bf31.1\\

    \bottomrule
    
    \end{tabular}
    \caption{Translation performance on WMT19/20 En$\Leftrightarrow$De. ``KD'' denotes the variant that exploits SNMT model to generate ST data with higher quality target.}
    \label{tab:target-quality}
\end{table}
\section{Related Work}
\paragraph{Unsupervised Neural Machine Translation} 
Before attempts to build NMT model using monolingual corpora only, unsupervised cross-lingual embedding mappings had been well studied by \citet{zhang-etal-2017-adversarial,artetxe2017learning,artetxe2018robust,conneau2017word}.
These methods try to align the word embedding spaces of two languages without parallel data and thus can be exploited for unsupervised word-by-word translation.
Initialized by the cross-lingual word embeddings, \citet{artetxe2018unsupervised} and \citet{lample2018unsupervised} concurrently proposed UNMT, which achieved remarkable performance for the first time using monolingual corpora only. Both of them rely on online back-translation and denoising auto-encoding. After that, \citet{lample2018phrase} proposed joint BPE for related languages and combined the neural and phrase-based methods. \citet{artetxe2019effective} warmed up the UNMT model by an improved statistical machine translation model. \citet{lample2019cross} proposed cross-lingual language model pretraining, which obtained large improvements over previous works. \citet{song2019mass} extended the pretraining framework to sequence-to-sequence. \citet{tran2020cross} induced data diversification in UNMT via cross-model back-translated distillation.

\paragraph{Data Augmentation} Back-translation~\cite{sennrich-etal-2016-improving,Edunov:2018:understanding,Marie:2020:ACL} and self-training~\cite{zhang2016exploiting,He:2020:revisiting,jiao-etal-2021-self} have been well studied in the supervised NMT. In the unsupervised scenario,  \citet{tran2020cross} have shown that multilingual pre-trained language models can be used to retrieve the pseudo parallel data from the large monolingual data. \citet{han2021unsupervised} use generative pre-training language models, e.g., GPT-3, to perform zero-shot translations and use the translations as few-shot prompts to sample a larger synthetic translations dataset. The most related work to ours is that offline self-training technology used to enhance low-resource UNMT~\citep{sun-etal-2021-self}. In this paper, the proposed online self-training method for UNMT can be applied to both high-resource and low-resource scenarios without extra computation to generate the pseudo parallel data.

\paragraph{Translationese Problem} Translationese problem has been investigated in machine translation evaluation~\cite{lembersky2012adapting,Zhang:2019:WMT,Edunov:2020:ACL,Graham:2020:EMNLP}. These works aim to analyze the effect of translationese in bidirectional test sets. In this work, we revisit the translationese problem in UNMT and find it causes the inaccuracy evaluation of UNMT performance since the training data entirely comes from the translated pseudo-parallel data.

\section{Conclusion}
Pseudo parallel corpus generated by back-translation is the foundation of UNMT. However, it also causes the problem of translationese and results in inaccuracy evaluation on UNMT performance. 
We attribute the problem to the data gap between training and inference and identify two data gaps, i.e., style gap and content gap. 
We conduct the experiments to evaluate the impact of the data gap on translation performance and propose the online self-training method to alleviate the data gap problems. 
Our experimental results on multiple language pairs show that the proposed method achieves consistent and significant improvement over the strong baseline XLM and MASS models on the test sets with natural input.  

\section*{Acknowledgements}
Zhiwei He and Rui Wang are with MT-Lab, Department of Computer Science and Engineering, School of Electronic Information and Electrical Engineering, and also with the MoE Key Lab
of Artificial Intelligence, AI Institute, Shanghai Jiao Tong University, Shanghai 200204, China.  Rui is supported by General Program of National Natural Science Foundation of China (6217020129), Shanghai Pujiang Program (21PJ1406800), and Shanghai Municipal Science and Technology Major Project (2021SHZDZX0102). Zhiwei is supported by CCF-Tencent Open Fund (RAGR20210119).

\bibliography{acl_latex}
\bibliographystyle{acl_natbib}

\appendix
\section{Training Details}
\label{sec:training-detail}
\subsection{Training Details of SNMT Model}
\label{sec:training-detail-snmt}

\paragraph{Training Data}
We use WMT16 parallel data for En-De and En-Ro and WMT14 for En-Fr. We randomly undersample the full parallel corpus.
The final sizes of En-De and En-Fr training corpus are 2M respectively, the size of En-Ro corpus is 400k.

\paragraph{Model}
We initialize the model parameter by XLM pre-trained model and adopt 2500 tokens/batch to train the SNMT model for 40 epochs. 
We select the best model by BLEU score on the validation set mentioned in~\cref{sec:experiemnts-setup}. 
Note that in order to avoid introducing other factors, our SNMT models are bidirectional, which is consistent with the UNMT models.

\subsection{Training Details of UNMT Model}
\label{sec:training-detail-unmt}

\paragraph{Training data}
\label{para:training-data-for-unmt}
\tablename~\ref{tab:data} lists the monolingual data used in this study to train the UNMT models\footnote{All the data is available at http://www.statmt.org/wmt20/translation-task.html except for En-De which we will release in our github repo.}.
We filter the training corpus based on language and remove sentences containing URLs.

\paragraph{Model} 
We adopt the pre-trained XLM models released by \citet{lample2019cross} and MASS models released by \citet{song2019mass} for all language pairs. 
In order to better reproduce the results for MASS on En-De, we use monolingual data to continue pre-training the MASS pre-trained model for 300 epochs and select the best model by perplexity (PPL) on the validation set. 
We adopt 2500 tokens/batch to train the UNMT model for 70 epochs and select the best model by BLEU score on the validation set. 

\paragraph{Hyper-parameter}
The target of self-training samples is the translation of the model, which may be noisy in comparison with the reference. Therefore, we adopted the strategy of linearly increasing $\lambda_S$ and keeping it at a small value to avoid negatively affecting the online back-translation training. We denote the beginning and final value of $\lambda_S$ by $\lambda^0_S$ and $\lambda^1_S$, respectively. We tune the $\lambda^0_S$ within $\{0, 1e{-3}, 1e{-2}, 2e{-2}\}$ and $\lambda^1_S$ within $\{5e{-3}, 5e{-2}, 1e{-1},1.5e{-1}\}$ based on the BLEU score on validation sets.

\begin{table}[htpb]
    \setlength\tabcolsep{3.2pt}
    \centering
    \begin{tabular}{c c r l}
    \toprule
    \bf Data & \bf Lang. & \bf \# Sent.  &   \bf Source\\
    \midrule
    \multirow{2}{*}{En-De}          & En & 50.0M  & \multirow{2}{*}{\citet{song2019mass}}          \\
                                    & De & 50.0M                                            \\
    \midrule
    \multirowcell{3}{En-Fr/Ro}  & En & 179.9M & \multirow{2}{*}{NC07-17}        \\
                                & Fr &  65.4M &    \\
                                & Ro &   2.8M & {NC07-17 + WMT16} \\
    \bottomrule
    \end{tabular}
    \caption{Data statistics for En-X translation tasks. ``M'' denotes millions. ``NC'' denotes News Crawl.}
    \label{tab:data}
\end{table}

\section{Sacrebleu Results}
\label{sec:sacrebleu-results}
To be consistent with previous works~\cite{lample2019cross,song2019mass,nguyen2021cbd}, we use \texttt{multi-bleu.perl} script in the main text to measure translation performance.
However, \citet{post-2018-call} has pointed out that \texttt{multi-bleu.perl}
requires user-supplied preprocessing, which cannot be directly compared and provide a sacreBLEU \footnote{https://github.com/mjpost/sacrebleu} tool to facilitate this.  
Although we adopted the same preprocessing steps for all models, we still report BLEU scores calculated with sacreBLEU \footnote{BLEU+c.mixed+\#.1+s.exp+tok.13a+v.1.5.1} in this section.
\cref{tab:snmt-unmt-translationese-sacrebleu,tab:natural-src-vs-translated-src-sacrebleu,tab:main-result-sacrebleu,tab:wmt19_20-sacrebleu,tab:target-quality-sacrebleu} show the sacreBLEU results of \cref{tab:snmt-unmt-translationese,tab:natural-src-vs-translated-src,tab:main-result,tab:wmt19_20,tab:target-quality}, respectively.

\begin{table*}
  \begin{minipage}{\columnwidth}
    \centering
    \hrule
    \resizebox{\columnwidth}{!}{%
    \begin{tabular}{c rr rr rr c}
        \toprule
        \multirow{2}{*}{\bf Model}& 
        \multicolumn{2}{c}{\bf En-Fr} & 
        \multicolumn{2}{c}{\bf En-De} &  
        \multicolumn{2}{c}{\bf En-Ro} &    
        \multirow{2}{*}{\bf Avg.}\\ 
        \cmidrule(lr){2-3} \cmidrule(lr){4-5} \cmidrule(lr){6-7} 
        & $\Rightarrow$~ & 
          $\Leftarrow$~  &  
          $\Rightarrow$~ & 
          $\Leftarrow$~  & 
          $\Rightarrow$~ & 
          $\Leftarrow$~ \\ 
          
    \midrule
    \multicolumn{8}{l}{\bf Full Test Set}\\
    SNMT & 37.3 & 33.4 & 29.7 & 33.8 & 33.8 & 32.4 & 33.4 \\
    \hdashline
    XLM   & 36.3 & 34.3 & 27.4 & 34.1 & 34.8 & 32.4 & 33.2 \\
    MASS  & 36.6 & 34.7 & 27.3 & 35.1 & 35.2 & 33.0 & 33.7 \\
    \midrule
    \multicolumn{8}{l}{\bf Target-Original Test Set / Translated Input}\\
    SNMT & 36.1 & 32.2 & 25.7 & 36.9 & 38.3 & 28.0 & 32.9 \\
    \hdashline
    XLM  & \bf37.8 & \bf36.2 & \bf26.9 & \bf42.0 & \bf42.2 & \bf34.1 & \bf36.5\\
    MASS & \bf37.9 & \bf37.3 & \bf27.3 & \bf42.7 & \bf43.2 & \bf35.2 & \bf37.3\\
    \midrule
    \multicolumn{8}{l}{\bf Source-Original Test Set / Natural Input}\\
    SNMT & \bf37.3 & \bf33.8 & \bf32.5 & \bf28.6 & \bf29.5 & \bf35.7 & \bf32.9 \\
    \hdashline
    XLM  & 33.8 & 30.2 & 26.8 & 22.5 & 27.6 & 30.2 & 28.5 \\
    MASS & 34.2 & 30.1 & 26.3 & 23.6 & 27.5 & 30.4 & 28.7  \\
    \bottomrule
    \end{tabular}%
    }
    \caption{SacreBLEU results of \tablename~\ref{tab:snmt-unmt-translationese}.}
    \label{tab:snmt-unmt-translationese-sacrebleu}
  \end{minipage}\hfill 
  \begin{minipage}{\columnwidth}
    \centering
    \hrule
    \resizebox{\columnwidth}{!}{%
    \begin{tabular}{c c c c c}
        \toprule
    \multirow{2}{*}{\bf Model} & \multicolumn{2}{c}{\bf Natural De} & \multicolumn{2}{c}{\bf Translated De$^{*}$} \\
    \cmidrule(lr){2-3} \cmidrule(lr){4-5}  
     & BLEU & $\Delta$ & BLEU & $\Delta$\\ 
    \midrule
    SNMT  & 28.6  & -- & 44.9 & -- \\
    \hdashline
    UNMT   & 22.5 & -6.1 & 42.0 & -2.9\\
    \bottomrule
    \end{tabular}%
    }
    \caption{SacreBLEU results of \tablename~\ref{tab:natural-src-vs-translated-src}.}
    \label{tab:natural-src-vs-translated-src-sacrebleu}

  \end{minipage}
\end{table*}

\begin{table*}[t]  
    \centering
    \begin{tabular}{l l l cc cc cc cc}
        \toprule
        \multirow{2}{*}{\bf Testset} &  \multirow{2}{*}{\bf Model} &    \multirow{2}{*}{\bf Approach}  &  
        \multicolumn{2}{c}{\bf En-Fr} & 
        \multicolumn{2}{c}{\bf En-De} &  
        \multicolumn{2}{c}{\bf En-Ro} &    
        \multirow{2}{*}{\bf Avg.}  &  \multirow{2}{*}{\bf $\Delta$}\\ 
        
        & & & $\Rightarrow$~ & 
          $\Leftarrow$~  &  
          $\Rightarrow$~ & 
          $\Leftarrow$~  & 
          $\Rightarrow$~ & 
          $\Leftarrow$~ \\ 

    \specialrule{.05em}{.1ex}{.1ex}
    \multicolumn{11}{c}{\textit{Our Implementation}}   \\
    \specialrule{.05em}{.1ex}{.1ex}
    \multirow{4}{*}{\bf Full set} & \multirow{2}{*}{XLM}    & UNMT    & 36.3 & 34.3 & 27.4 & 34.1 & 34.8 & 32.4 & 33.2 & -- \\

    & & ~~+Self-training                                              & \bf36.7 & \bf34.9 & \bf28.3 & \bf34.6 & \bf36.3 & \bf33.7 & \bf34.1 & +0.9\\

     \cline{2-11}
    &  \multirow{2}{*}{MASS}    & UNMT                                & 36.6 & 34.7 & 27.3 & 35.1 & 35.2 & 33.0 & 33.7 & --\\

     & & ~~+Self-training                                             & \bf36.8 & \bf35.0 & \bf29.1 & \bf35.5 & \bf36.6 & \bf33.7 & \bf34.4 & +0.7\\

    \midrule
    \multirow{4}{*}{\bf Trg-Ori}  &  \multirow{2}{*}{XLM}  & UNMT     & 37.8 & 36.2 & \bf26.9 & 42.0 & 42.2 & \bf34.1 & 36.5 & --\\

    & & ~~+Self-training                                              & \bf38.0 & \bf37.5 & 26.7 & \bf42.1 & \bf42.9 & 33.8 & \bf36.8 & +0.3\\

     \cline{2-11}
     &  \multirow{2}{*}{MASS}                             & UNMT      & \bf37.9 & \bf37.3 & 27.3 & \bf42.7 & \bf43.2 & \bf35.2 & \bf37.3 & --\\

     & & ~~+Self-training                                             & 37.7 & 37.0 & \bf27.9 & 42.5 & 43.0 & 34.9 & 37.2 & -0.1\\

    \midrule
    \multirow{4}{*}{\bf Src-Ori}  &  \multirow{2}{*}{XLM}  & UNMT     & 33.8 & \bf30.2 & 26.8 & 22.5 & 27.6 & 30.2 & 28.5 & --\\

    & & ~~+Self-training                                              & \bf34.4 & 30.1 & \bf28.2 & \bf23.2 & \bf29.7 & \bf32.4 & \bf29.7 & +1.2\\

     \cline{2-11}
    &  \multirow{2}{*}{MASS}    & UNMT                                & 34.2 & 30.1 & 26.3 & 23.6 & 27.5 & 30.4 & 28.7 & --\\

    & & ~~+Self-training                                              & \bf34.9 & \bf30.7 & \bf28.9 & \bf24.9 & \bf30.3 & \bf31.5 & \bf30.2 & +1.5\\

    \bottomrule
    \end{tabular}
    \caption{SacreBLEU results of \tablename~\ref{tab:main-result}.}
    \label{tab:main-result-sacrebleu}
\end{table*}

\begin{table}[htpb]   
\centering
\setlength\tabcolsep{3.3pt}
    \begin{tabular}{l ll  ll cc}
        \toprule
        \bf Model & 
        \multicolumn{2}{c}{\bf WMT19} & 
        \multicolumn{2}{c}{\bf WMT20} & 
        \multirow{2}{*}{\bf Avg.}    &  \multirow{2}{*}{\bf $\Delta$} \\
    
        \cmidrule(lr){2-3} \cmidrule(lr){4-5}
    
        \bf{~~+Approach} & \bf ~~$\Rightarrow$ & 
          \bf ~~$\Leftarrow$  &  
          \bf ~~$\Rightarrow$ & 
          \bf ~~$\Leftarrow$  \\ 
        \midrule

   XLM \\
     ~~+UNMT & 25.8 & 24.1 & 21.8 & 26.3 & 24.5 & -- \\
     ~~+Offline ST               & 26.0 & 23.9 & 22.0 & 25.8 & 24.4 & -0.1 \\
     ~~+CBD                      & \bf27.4 & 25.2 & \bf23.0 & 26.7 & 25.6 & +1.1\\
     ~~+Online ST                & \bf27.4 & \bf25.8 & 22.8 & \bf27.1 & \bf25.8 & +1.3 \\
    \midrule
   MASS \\
     ~~+UNMT & 26.0 & 24.3 & 22.1 & 26.5 & 24.7 & --\\
     ~~+Offline ST              & 26.4 & 24.2 & 22.1 & 26.4 & 24.8 & +0.1\\
     ~~+CBD                     & 27.4 & 25.2 & \bf22.9 & 26.6 & 25.5 & +0.8 \\
     ~~+Online ST               & \bf27.7 & \bf25.7 & 22.8 & \bf27.4 & \bf25.9 & +1.2 \\
    \bottomrule
    \end{tabular}
    \caption{SacreBLEU results of \tablename~\ref{tab:wmt19_20}.}
    \label{tab:wmt19_20-sacrebleu}
\end{table}

\begin{table}[htpb]
    \centering
    \begin{tabular}{c cc cc}
    \toprule
    \multirow{2}{*}{\bf Approach} & \multicolumn{2}{c}{\bf WMT19} & \multicolumn{2}{c}{\bf WMT20}  \\
    \cmidrule(lr){2-3} \cmidrule(lr){4-5}

    & \bf ~$\Rightarrow$ & 
      \bf ~$\Leftarrow$  &  
      \bf ~$\Rightarrow$ & 
      \bf ~$\Leftarrow$  \\ 
      
     \midrule
        \multicolumn{5}{c}{\bf XLM} \\
    UNMT   & 25.8 & 24.1 & 21.8 & 26.3 \\
    ~+ST   & 27.4 & 25.8 & 22.8 & 27.1 \\
    ~~+KD  & \bf32.4 & \bf30.6 & \bf27.9 & \bf29.7 \\
    \hdashline
    \multicolumn{5}{c}{\bf MASS} \\
    UNMT   & 26.0 & 24.3 & 22.1 & 26.5\\
    ~+ST   & 27.7 & 25.7 & 22.8 & 27.4\\
    ~~+KD  & \bf31.8 & \bf30.5 & \bf30.1 & \bf30.6\\

    \bottomrule
    
    \end{tabular}
    \
    \caption{SacreBLEU results of \tablename~\ref{tab:target-quality}.}
    \label{tab:target-quality-sacrebleu}
\end{table}

\section{Translation Examples}
\label{sec:tranlstion_examples}
\begin{table*}[htpb]
    \centering
    \begin{tabular}{l l}
    \toprule
    Source & Mindestens ein \textcolor{red}{Bayern-Fan} wurde verletzt aus dem Stadion transportiert . \\
    Reference & At least one \textcolor{red}{Bayern} fan was taken injured from the stadium .\\
    UNMT    &  At least one  \textcolor{blue}{Scotland fan} was transported injured from the stadium .\\
                            
    \midrule
    Source & Übrigens : \textcolor{red}{München} liegt hier ausnahmsweise mal nicht an der Spitze . \\

    Reference & Incidentally , for once \textcolor{red}{Munich} is not in the lead .\\
    
    UNMT    &  Remember , \textcolor{blue}{Edinburgh} is not at the top of the list here for once .\\
        \midrule

    Source & Justin Bieber in der Hauptstadt : Auf Bieber-Expedition in \textcolor{red}{Berlin} \\
  
    Reference & Justin Bieber in the capital city : on a Bieber expedition in \textcolor{red}{Berlin}\\
   
    UNMT    &  Justin Bieber in the capital : On Bieber-inspired expedition in  \textcolor{blue}{NYC}\\
        \midrule

    Source & Zum Vergleich : In diesem Jahr werden in \textcolor{red}{Deutschland} 260.000 Einheiten fertig . \\
   
    Reference &In comparison , 260,000 units were completed in this year in \textcolor{red}{Germany}.\\
    
    UNMT    &  To date , 260,000 units are expected to be finished in the  \textcolor{blue}{UK} this year .\\
        \midrule

    \multirow{2}{*}{Source} & \textcolor{red}{Deutschland} schiebe ein Wohnungsdefizit vor sich her , das von Jahr zu Jahr \\
                            & größer wird . \\
    Reference & \textcolor{red}{Germany} has a housing deficit which increases every year .\\
    \multirow{2}{*}{UNMT}    &  The  \textcolor{blue}{U.S.} was shooting ahead of a housing deficit that is expected to grow from year \\
            & to year .\\
    \bottomrule
    \end{tabular}
    \caption{Example translations in WMT16 De$\Rightarrow$En. the UNMT model outputs the hallucinated translations which are biased towards the target language En.}
    \label{tab:transltation_examples}
\end{table*}
Table~\ref{tab:transltation_examples} presents several example translations that the UNMT model outputs the hallucinated translations, which are biased towards the target language.

\end{document}